\documentclass[pmlr,twocolumn,10pt]{jmlr} 

\usepackage{booktabs}
\usepackage{siunitx}
\usepackage[switch]{lineno}
\usepackage{makecell}
\usepackage{array}
\usepackage{hyperref}
\usepackage{algorithm}
\usepackage{caption}

\jmlrvolume{287}
\jmlryear{2025}
\jmlrsubmitted{} 
\jmlrpublished{} 
\jmlrworkshop{Conference on Health, Inference, and Learning (CHIL) 2025}

\title[KEEP: Knowledge-preserving and Empirically refined Embedding Process]{KEEP: Integrating Medical Ontologies with Clinical Data for Robust Code Embeddings}

\newcommand{\addrColumbia}{1}
\newcommand{\addrNYGC}{2}

\author{%
\Name{Ahmed Elhussein}$^{\addrColumbia, \addrNYGC}$\Email{ae2722@cumc.columbia.edu}\\
\Name{Paul Meddeb}$^{\addrNYGC}$\Email{paul.meddeb@etu.minesparis.psl.eu}\\
\Name{Abigail Newbury}$^{\addrColumbia, \addrNYGC}$\Email{abigail.newbury@columbia.edu}\\
\Name{Jeanne Mirone}$^{\addrNYGC}$\Email{ jeanne.mirone@etu.minesparis.psl.eu}\\
\Name{Martin Stoll}$^{\addrNYGC}$\Email{martin.stoll@etu.minesparis.psl.eu}\\
\Name{Gamze G{\"u}rsoy}$^{\addrColumbia, \addrNYGC}$ \Email{gamze.gursoy@columbia.edu}\\
\addr$^{\addrColumbia}$Columbia University, USA\\
\addr$^{\addrNYGC}$New York Genome Center, USA
}

\begin{document}
\maketitle
\begin{abstract}
Machine learning in healthcare requires effective representation of structured medical codes, but current methods face a trade-off: knowledge graph-based approaches capture formal relationships but miss real-world patterns, while data-driven methods learn empirical associations but often overlook structured knowledge in medical terminologies. We present KEEP (Knowledge-preserving and Empirically refined Embedding Process), an efficient framework that bridges this gap by combining knowledge graph embeddings with adaptive learning from clinical data. KEEP first generates embeddings from knowledge graphs, then employs regularized training on patient records to adaptively integrate empirical patterns while preserving ontological relationships. Importantly, KEEP produces final embeddings without task-specific axillary or end-to-end training enabling KEEP to support multiple downstream applications and model architectures. Evaluations on structured EHR from UK Biobank and MIMIC-IV demonstrate that KEEP outperforms both traditional and Language Model-based approaches in capturing semantic relationships and predicting clinical outcomes. Moreover, KEEP’s minimal computational requirements make it particularly suitable for resource-constrained environments.
\end{abstract}

\paragraph*{Data and Code Availability}
This research has been conducted using data from UK Biobank \citep{sudlow2015uk} and MIMIC-IV \cite{johnson2021mimic}. Researchers can request access via \url{https://www.ukbiobank.ac.uk/} and \url{https://physionet.org/content/mimiciv/3.1/}, respectively. Implementation code is available at \url{https://github.com/G2Lab/keep}

\paragraph*{Institutional Review Board (IRB)}
This study does not require IRB approval as all data used are publicly
available.

\section{Introduction}
\label{sec:intro}
Structured electronic health records (EHRs) contain extensive data across multiple domains including diagnoses, medications, procedures, and clinical observations. These datasets are a major resource for developing machine learning (ML) models across various healthcare applications including predictive modeling, phenotyping, and drug repurposing \citep{tang2024harnessing}. However, the discrete nature of medical codes within these datasets presents challenges for effective representation learning in ML. Traditional discrete representations like one-hot encoding produce extremely high-dimensional and sparse representations \citep{johnson2021encoding}. These high-dimensional, sparse representations create several challenges for gradient descent algorithms: they slow convergence rates, dilute the strength of relevant signals, and ultimately restrict the model's ability to generalize effectively \citep{guo2016entity}. Additionally, one-hot encoding treats each code as independent, ignoring the relationships defined in their source medical terminologies. However, these terminologies contain knowledge graphs that encode essential clinical and biological relationships. By failing to capture these relationships, one-hot encoding discards valuable context that is essential for more sophisticated tasks \citep{chang2020benchmark, choi2016multi}.

Representation learning addresses these limitations by transforming high-dimensional medical data into embeddings, which are compact real-valued vector representations \citep{bengio2013representation}.  Research in Natural Language Processing (NLP) has shown that sequence-based representation learning can transform high dimensional discrete data into dense vector spaces that capture semantic relationships through distance metrics \citep{mikolov2013efficient, pennington2014glove}. In healthcare applications, ML models using real-valued embeddings have demonstrated superior performance in tasks such as risk prediction and patient stratification \citep{choi2018mime, rasmy2021med}.

Recently, language models (LMs) have been used for medical code representation, offering several advantages. First, unlike traditional methods that produce static embeddings, LMs generate dynamic, context-aware representations that adapt to the surrounding information. This capability allows them to capture nuanced relationships between medical codes more effectively \citep{alsentzer2019publicly, pang2021cehr, rasmy2021med}. Second, these models benefit from pre-training on extensive unstructured biomedical corpora, including PubMed and clinical notes, allowing them to learn valuable domain knowledge from many sources \citep{lee2020biobert, luo2022biogpt}. Third, LMs excel at transfer learning, efficiently adapting to new tasks with limited data through their few-shot learning capabilities \citep{brown2020language}. However, applying LMs to structured medical data also has limitations. The subword tokenization approach fragments the unit meanings of medical codes, introducing semantic ambiguity and failing to preserve hierarchical structures \citep{dwivedi2024representation, yuan2022coder}. This fragmentation is particularly problematic because medical codes are designed as complete, atomic units with precise meanings. LMs also lack mechanisms to directly integrate knowledge graphs from medical ontologies, resulting in embeddings that inadequately capture known biomedical relationships \citep{chang2024use}. The practical impact of these limitations is clear: without fine-tuning, even GPT-4 achieves less than 50\% accuracy on basic medical code matching tasks \citep{soroush2024large}. While fine-tuning can partially address these challenges, it demands labeled data and substantial computational resources \citep{zhang2024scaling}.

Prior work in medical code representation reveals three critical requirements: (1) capturing semantic relationships from medical knowledge graphs, (2) reflecting empirical patterns observed in patient records, and (3) demonstrating robust utility across diverse healthcare tasks \citep{si2021deep}. We propose KEEP (Knowledge-preserving and Empirically refined Embedding Process), an efficient framework that addresses these requirements through a two-step approach. First, KEEP utilizes knowledge graphs to generate initial embeddings that preserve semantic relationships. Next, it refines these embeddings through regularized training on empirical co-occurrence data, ensuring rare codes retain meaningful ontological representations while frequently co-occurring codes benefit from data-driven adjustments. With its computational efficiency and compatibility with existing data formats, KEEP enables broad generalizability across healthcare settings. Comprehensive empirical evaluations demonstrate KEEP's robust performance, consistently outperforming traditional and LM-based models in both intrinsic tasks (semantic relationship encoding) and extrinsic tasks (clinical prediction).

Our framework complements recent advances in LMs in two key ways by enabling more effective bridging between structured medical knowledge and unstructured data. First, it generates structured data representations that can be integrated with LM-derived embeddings through multimodal integration techniques \citep{ebrahimi2023lanistr}. Second, it could enhance LM performance on medical codes by providing domain-aware initialization for fine-tuning, enabling more effective incorporation of medical ontologies and relationships \citep{hewitt2021initializing, fatemi2023talk}.

\section{Related Works}
Early efforts to create medical code embeddings built directly on advances in NLP, treating medical codes as tokens and patient records as sequences \citep{de2014medical}. This approach evolved with Med2Vec, which incorporated visit-level information to better capture co-occurrence patterns \citep{choi2016multi}. However, recognizing that medical codes differ fundamentally from natural language through their well-defined ontological structure, researchers developed approaches that leverage knowledge graphs to generate more semantically meaningful embeddings \citep{grover2016node2vec, agarwal2019snomed2vec}. \citet{yuan2022coder} further advanced this direction by incorporating medical knowledge graphs through contrastive learning techniques.

The development of LMs marked a significant shift in medical code representation. BERT-based models such as BioBERT and ClinicalBERT introduced context-dependent embeddings and leveraged pre-training on unstructured biomedical corpora like PubMed and MIMIC-III \citep{alsentzer2019publicly, lee2020biobert}. These innovations significantly improved performance on tasks like named entity recognition and relation extraction. However, their application to structured data was constrained by tokenization strategies ill-suited for medical codes. Models like Med-BERT and CEHR-BERT tried to address this limitation by introducing visit-level tokenization tailored for structured EHR data \citep{rasmy2021med, pang2021cehr}. Recent work has introduced purpose-built foundation models specifically designed for structured longitudinal health data \citep{dwivedi2024representation, steinberg2021language}. Alternatives to code tokenization have explored generating embeddings by using descriptions rather than the code itself, achieving notable gains in predictive performance \citep{kane2023compressed, lin2020patient, lee2024emergency}. 

Despite these advancements, significant challenges remain. The non-Identically Independently Distributed (IID) nature of medical data necessitates institution-specific model fine-tuning for pre-trained model, requiring substantial computational resources and introducing deployment complexity. This presents a significant barrier to widespread implementation, particularly for institutions with limited resources \citep{wornow2023shaky}. Practical implementation is further hindered by the fact that embeddings from recent structured EHR models are rarely made publicly available, which limits external validation of the models \citep{wornow2023shaky}. Current methods also struggle to fully capture the hierarchical nature of medical coding systems, where codes exist in complex parent-child relationships that influence their semantic meaning. 

\section{Background}
\subsection{OMOP knowledge graph}
\label{background_omop}
The Observational Medical Outcomes Partnership (OMOP) Common Data Model (CDM) is a standardized data model that transforms heterogeneous structured EHR data into a consistent format. Like most medical terminologies, OMOP maintains an associated knowledge graph that captures relationships between clinical concepts through various semantic connections. The graph structure encodes both hierarchical relationships (\emph{e.g.,} "Type 2 Diabetes" is-a "Diabetes Mellitus") and non-hierarchical clinical associations (\emph{e.g.,} "may-prevent", "has-finding-site"). These relationships provide valuable domain knowledge but are limited to pre-defined medical associations. They do not capture the full complexity of relationships or empirically observed patterns, which can vary across healthcare settings.

\subsection{Resnik similarity}
Resnik similarity measures semantic similarity between two nodes in a graph by evaluating the information content (IC) of their least common ancestor (LCA). For any concept $c$, IC is defined as $IC(c):=-\log p(c)$, where $p(c)$ represents the proportion of nodes that are descendants of $c$ in the hierarchy. This formulation assigns higher IC to specific, rare concepts and lower IC to common, general concepts. Nodes sharing a more specific LCA are considered more semantically similar.

\subsection{Models}
Below we discuss two complementary approaches to embedding construction leveraged by our framework. We also discuss LMs to contextualize current approaches in medical code representation.

\textbf{GloVe} (Global Vectors for Word Representation) learns semantic relationships by analyzing co-occurrence patterns in data \citep{pennington2014glove}. It uses a co-occurrence matrix $X$, where $X_{ij}$ represents how often concept $i$ appears with concept $j$ to minimize:
\begin{equation}
\sum_{i,j=1}^{V} f(X_{ij}) \left( w_i^\top \tilde{w}_j - \log X_{ij} \right)^2,
\end{equation}
where $V$ denotes vocabulary size, $w_i$ and $\tilde{w}_j$ are the learned concept vectors, and $f(X_{ij})$ weights frequent co-occurrences to prevent them from dominating the optimization.

\textbf{node2Vec} generates embeddings that preserve graph structure by simulating biased random walks through a knowledge graph \citep{grover2016node2vec}. For each node $u$, it maximizes the probability of observing its network neighborhood $\mathcal{N}_S(u)$ given its embedding $f(u)$:
\begin{equation}
\max_f \sum_{u \in V} \log \Pr(\mathcal{N}_S(u) \mid f(u)),
\end{equation}
where $V$ represents the set of nodes, $f(u)$ maps node $u$ to a vector in $\mathbb{R}^d$ and $\Pr(v \mid f(u))$ is the probability of node $v$ being sampled as part of $\mathcal{N}_S(u)$ given $f(u)$. Assuming conditional independence, this probability factorizes as:
\begin{equation}
\Pr(\mathcal{N}_S(u) \mid f(u)) = \prod_{v \in \mathcal{N}_S(u)} \Pr(v \mid f(u)).
\end{equation}

\textbf{Language Models (LMs)} are the current state-of-the-art in text representation but face limitations with structured medical data. BERT uses bidirectional transformers to generate context-aware embeddings \citep{kenton2019bert}. GPT models process text unidirectionally and with embeddings usually averaged from a decoder layer in the absence of a [CLS] token \citep{achiam2023gpt}.

\subsection{Code representation}
\label{background_code-rep}
Recent advances in medical code representation have been dominated by data-driven approaches, particularly foundation models that employ autoregressive techniques like masked token prediction. Despite their success in general language tasks, these models face distinct challenges in the medical domain. They struggle to capture the explicit hierarchical relationships within medical ontologies and often generate poor representations of rare conditions due to limited clinical examples \citep{banerjee2023machine}. Furthermore, by learning from individual patient sequences in isolation, these models cannot explicitly leverage patterns shared across patients. Both limitations increase the risk of overfitting to institution-specific data. This dependence could compromise their generalizability across different healthcare settings.

Knowledge graphs offer an alternative foundation for medical code representation by explicitly encoding relationships between clinical concepts \citep{lee2021comparative}. These structured graphs preserve semantic relationships and enable models to incorporate established medical knowledge. However, a major limitation of knowledge graphs is their inability to comprehensively capture the full complexity of biomedical relationships. Even the most detailed knowledge graphs have fundamental limitations: they cannot encode all relationships, especially those that emerge from real-world data, and their encoded relationships fail to capture the full nature of relationships, particularly the context-dependence (see Section \ref{background_omop}). These inherent constraints mean that representations based solely on knowledge graphs fail to reflect all associations observed in practice.

\section{Methods}
\subsection{Problem setup}
Our objective is to develop medical code embeddings that integrate structured knowledge graphs with real-world medical data. This approach captures both the taxonomic design of medical codes and their practical clinical usage, enabling the embeddings to reflect biological relationships and real-world comorbidity patterns. We believe such integrated representations can enhance downstream analytical tasks. Additionally, we aim to ensure these embeddings can be generated with minimal computational requirements, making them accessible to a wide range of institutions.

\subsection{Data}
In our analysis, we use the UK Biobank (UKBB) dataset \citep{papez2023transforming}, comprising structured outpatient and hospital records for approximately 500,000 patients and the MIMIC-IV dataset comprising of over 200,000 patients admitted to the ICU. Both datasets were converted to the OMOP Common Data Model. We used OMOP CDM due to its richer graph structure and multi-domain integration. Note, however, that any terminology with an associated knowledge graph can be used (\emph{e.g.,} SNOMED, ICD) instead.

To generate embeddings, we require two key components derived from the EHR and their associated terminology: a knowledge graph and a co-occurrence matrix. We construct the OMOP knowledge graph using only condition concepts connected through hierarchical 'is-a' relationships. To maintain computational efficiency and focus on useful concept abstractions, we limit the hierarchical depth of our knowledge graph to five levels from the root node. Codes further away from this are aggregated to their parent node (see Section \ref{app:graph} for more details). We found that a depth of 5 provided a level of specificity comparable to ICD codes, making it well-suited for our task. While our current implementation utilizes only 'is-a' relationships, the framework can accommodate more complex relationships and additional concept types, such as drugs and procedures, enabling multi-domain knowledge integration as enriching the knowledge graph has been shown to improve embedding quality \citep{shen2019hpo2vec+}.

For the co-occurrence matrix, we analyze complete patient histories rather than individual visits to reduce matrix sparsity, though visit-level analysis could provide more granular comorbidity relationships given sufficient data volume. To ensure reliable phenotyping, we require two occurrences of a diagnosis code in a patient's record to label a patient with that condition (see Section \ref{app:cooc_mat} for more details).

\subsection{Algorithm}
\begin{figure*}
    \centering
    \includegraphics[width=0.95\linewidth]{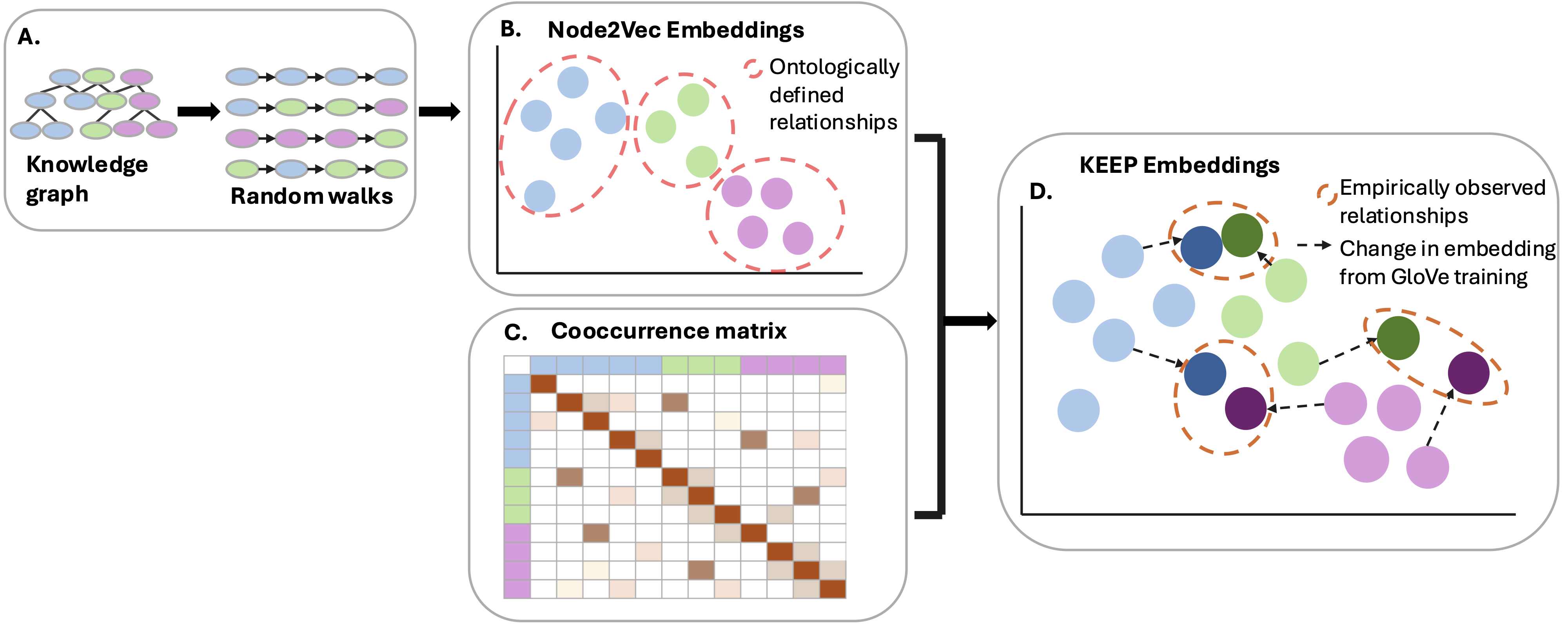}
    \caption{Overview of KEEP's approach: (A) Generate random walks on knowledge graph. (B) Walks used to create initial embeddings whose dimensions align with the ontology. (C) A co-occurrence matrix is constructed from EHR data. (D) GloVe model is initialized with the embeddings from (B) and regularized to incorporate empirical relationships from (C) while preserving ontologically-aligned dimensions. That is embeddings are adjusted based on the strength of observed associations.}
        \label{fig:keep_overview}
\end{figure*}

Our method, KEEP, is summarized in Figure \ref{fig:keep_overview} and Algorithm \ref{alg:KEEP_pseudocode}. KEEP addresses the limitations discussed in Section \ref{background_code-rep} by integrating knowledge graphs with real-world clinical data. This integration creates embeddings that are both theoretically grounded and empirically validated, promoting robustness and generalizability across institutions. KEEP takes a more intuitive approach to integrating knowledge graphs with EHR data. Rather than treating graph knowledge as an additional modeled input, we approach it as a prior that regularizes learning from real-world data—more closely mimicking how physicians learn. Further, by prioritizing computational efficiency, we aim to support representation learning even in resource constrained hospitals.

KEEP follows a two-stage training procedure that combines established lightweight algorithms through regularized training. The first stage employs node2vec to generate initial embeddings based on a knowledge graph. This step, independent of institution-specific data, captures the fundamental biological relationships encoded in medical ontologies. Recent research has demonstrated that these knowledge graph-based embeddings alone effectively capture phenotype-aligned relationships \citep{lee2021comparative}.

In the second stage, KEEP uses the node2vec embeddings as initialization values for a modified GloVe model. To preserve the valuable medical knowledge while incorporating empirical patterns, we augment the standard GloVe objective function with a regularization term. This approach prevents catastrophic forgetting during training \citep{kirkpatrick2017overcoming}, ensuring the final embeddings maintain their ontological foundation while incorporating real-world associations. In our experiments, both node2vec and GloVe models produced embeddings of identical dimensionality. While maintaining consistent dimensions between both models is recommended, if working with pretrained embeddings of different dimensions, we suggest using autoencoders to either reduce or expand the node2vec embedding dimensions to match those required by the GloVe model.

The objective function for KEEP combines the GloVe loss with a regularization term:
\begin{equation}
\underbrace{\sum_{i,j=1}^{V} f(X_{ij}) \left( w_i^\top \tilde{w}_j - \log X_{ij} \right)^2}_{{\text{GloVe}}}
+
\underbrace{\lambda \sum_{i=1}^{V} | w_i - w_i^{\text{n2v}} |^2}_{\text{Regularization Term}}
\end{equation}

Here, $\lambda$ controls the regularization strength and $w_i^{\text{n2v}}$ represents the initial node2vec embedding for code $i$. This formulation provides two key technical advantages. It maintains robust representations for rare codes by anchoring their embeddings to biologically meaningful dimensions established through the node2vec initialization. Additionally, the regularization term enables controlled adaptation of the knowledge graph-based initialization as more comorbidity data becomes available, with $\lambda$ determining the degree of permissible deviation. This approach effectively creates a continuum between purely knowledge-based and purely data-driven representations, with $\lambda$ serving as the control parameter.

\section{Experiment setup}
We evaluate KEEP's embeddings against several baseline techniques through both intrinsic and extrinsic assessments. The intrinsic evaluation examines the relationships captured within the embedding space, while the extrinsic evaluation measures their effectiveness in downstream clinical prediction tasks.

\subsection{Comparator embeddings}
\label{subsec:comparator_embeddings}
We evaluate KEEP's embeddings against two categories of existing approaches: pretrained language model embeddings and specialized embeddings trained on this data. This allows us to assess the relative advantages of our method across different use cases and computational requirements.

\subsubsection{Pre-trained Language Model Embeddings}
We generated embeddings from pre-trained LMs using OMOP code descriptions, following the methodology of \citet{kane2023compressed}. For BERT architectures, we extracted the \texttt{[CLS]} token from the final layer. For BioGPT, we used the mean of the final hidden state, as this can capture semantics in a continuous latent space \citep{Hao2024-trainllm}.

We evaluated three leading pre-trained models for medical tasks: BioClinBERT \citep{alsentzer2019publicly}, fine-tuned on MIMIC clinical notes after PubMed training; ClinBERT \citep{wang2023optimized}, trained on extensive EHR data from over 3 million patients; and BioGPT \citep{luo2022biogpt}, a transformer model trained on PubMed biomedical literature.

To capture hierarchical relationships, we generated embeddings that incorporated that also include the descriptions of parent and grandparent codes. Hierarchical information was explicitly encoded by appending 'is a' followed by the description. This provided the models with a contextual understanding of the taxonomic structure when generating embeddings. This is a similar approach as proposed by \citet{deng2024graphvis}. We refer to these embeddings as Hierarchy Aware (HA). For more details see Section \ref{app:pre_trained_emb}

\subsubsection{Dataset-Specific Embeddings}
We evaluated several embedding approaches that derive representations directly from structured medical data. To establish baselines for KEEP, we separately implemented node2vec and GloVe (see Sections \ref{app:n2v} and \ref{app:glove} for training details). We included Cui2Vec \citep{beam2020clinical} in our comparison due to its demonstrated ability to capture semantic relationships and enhance performance across various downstream tasks. Additionally, we also added a Graph Attention Network (GAT) baseline, following the approach of \citep{piya2024healthgat}, with Node2Vec-based initialization and a co-occurrence-based graph construction. However, our implementation differs from the original HealthGAT in two key aspects. First, we do not integrate a supervised auxiliary task during embedding generation, as this is designated for extrinsic evaluation. Second, we use only diagnoses data, excluding procedures.

\subsection{Assessing the Impact of Regularization}
We first assess KEEP's ability to effectively combine knowledge graph structure with co-occurrence patterns. Our analysis aims to verify that KEEP effectively combines both data sources, demonstrating stronger co-occurrence relationships than node2vec and better graph relationships than GloVe. While not direct measures of quality, these assessments help validate the effective integration of both training objectives in our embeddings.

We measured the correlation between embedding cosine similarities and co-occurrence values to assess preservation of empirical relationships (See Section \ref{app:impact_reg}). For hierarchical relationships, we computed correlations between embedding cosine similarities and Resnik similarity scores between node pairs. Comparing these metrics across different embedding approaches provides insight into how each method balances graph-based and empirical relationships. Note, we only used UK Biobank dataset that contains a more diverse range of diseases.

\subsection{Intrinsic Evaluation}
Our intrinsic evaluation framework assesses how effectively embeddings capture known relationships. Following the methodology of \citep{beam2020clinical}, we focus on discriminating between genuine  relationships and random associations. To do this, we curated a set of test codes that span both data-driven relationships (\emph{e.g.,} obesity) and medical relationships derived from medical ontologies (\emph{e.g.,} type 1 diabetes) (see Table \ref{tab:core_concepts_comorbidities} for list of concepts). For each disease, we evaluate its representation by comparing the cosine similarity between its embedding and those of its known condition set against a null distribution generated from similarity scores with randomly selected conditions. A relationship is defined as correctly identified if its similarity score exceeds the 95th percentile of this disease-specific null distribution. Given our analysis spans 6 diseases, we employ the Wilcoxon signed-rank test to assess systematic differences in identification performance across embedding methods. Note, for this task we only used UK Biobank dataset that contains a more diverse range of diseases. For a full description of the method see Section \ref{app:impact_reg}.

\subsection{Extrinsic Evaluation}
\subsubsection{Clinical tasks}
We evaluated our embeddings' practical utility through clinical prediction tasks focused on two distinct datasets—the UK Biobank (UKBB) and MIMIC-IV - focusing on clinically diverse and consequential outcomes. In the UKBB cohort we evaluated rare complications in chronic disease populations (see Table \ref{tab:complications_prevalence}). Our prediction tasks examined: diabetic sequelae (including eye complications, peripheral vascular disease, neuropathy, and kidney disease) in patients with type 2 diabetes; myocardial infarction in patients with angina; and acute renal failure in patients with chronic kidney disease (see Table \ref{tab:outcome_codes} for list of codes used). These complications were selected as improved risk stratification in these areas can significantly impact patient care decisions \citep{ndjaboue2022prediction,ramspek2020towards,reeh2019prediction}. Additionally, these complications involve the interaction of multiple diseases, providing a robust framework for evaluating the ability of embeddings to capture complex relationships. Complementing this, we analyzed 30-day readmission/representation risk in the MIMIC-IV dataset for patients discharged directly from emergency departments. By spanning chronic disease complications (UKBB) and acute care transitions (MIMIC-IV), our evaluation framework rigorously probes embedding capabilities.

 \begin{table}[ht]
\centering
\caption{Prevalence of Complications (UK Biobank) and Readmission (MIMIC-IV)}
\begin{tabular}{lcc}
\toprule
\textbf{Cohort} & \textbf{Complications} & \textbf{Prevalence} \\ \Xhline{1pt}
Type 2 DM & All Diabetes & 17.2\% \\
Type 2 DM& Peripheral vascular & 1.7\% \\
Type 2 DM & Kidney & 2.8\% \\
Type 2 DM & Eye & 9.7\% \\
Type 2 DM & Neuropathy & 3.4\% \\
Angina & MI & 9.6\% \\
CKD & Renal failure & 7.0\% \\
ER discharge & Readmission & 3.0\% \\
\bottomrule
\end{tabular}
\label{tab:complications_prevalence}
\end{table}

\subsubsection{Model training and evaluation}
For our prediction tasks, we implemented a single-layer encoder model with 4 attention heads to process sequences of patient diagnoses. The final representation is passed through a feedforward layer for prediction. Each diagnosis code was represented using the various embedding approaches under evaluation. We retained the original embedding dimensions for each method to preserve information content, though this resulted in varying model sizes, particularly for the pre-trained LM embeddings which typically have larger dimensions (see Table \ref{tab:model_embeddings_dim} for dimensions). For the trained embeddings, we use a fixed dimension size of 100 as preliminary analysis using the reconstruction losses for GloVe and cui2vec models showed a plateau at 100 dimensions. Given that the pre-trained models have fixed dimensions, we standardized embedding dimensions to ensure fair performance comparison.

We conducted a learning rate sweep across evenly spaced values ranging from $1e{-1}$ to $5e^{-5}$, using the AdamW optimizer for all experiments. For each learning rate, we performed five training runs and selected the learning rate yielding the lowest median loss. Model performance was evaluated through 100 independent runs, with bootstrap sampling across these runs to generate 95\% confidence intervals. See Section \ref{app:ext_eval} for full training details.

For model evaluation, we use Test loss, AUPRC, AUC, Mathews Correlation Coefficient (MCC), and F1 score. To make generalizable statements across all tasks, we evaluated differences in the overall rank between embeddings using the Wilcoxon signed-rank test. AUPRC is presented in the main manuscript, rest are found at Section \ref{app:exp_results}.

\section{Experiment Results}
\subsection{Assessing the Impact of Regularization}
Our combined method effectively balanced both co-occurrence and hierarchical relationships, demonstrating strong performance on both metrics. Specifically, we improved upon node2vec's co-occurrence correlation and surpassed GloVe's Resnik similarity correlation. These results validate our approach's ability to preserve both graph-based structure and empirical relationships in the final embeddings.

Notably, enriching language model inputs with hierarchical descriptions improved their ability to capture taxonomic relationships without compromising co-occurrence pattern recognition. However, language models showed substantially lower co-occurrence similarity compared to embeddings trained directly on the dataset. This performance discrepancy likely stems from differences in training data sources—our embeddings were trained on UK health system data, while the language models were trained on US intensive care and Chinese healthcare datasets. These geographic and healthcare system variations influence the captured relationships, reflecting distinct clinical practices, coding patterns, and patient populations.

\begin{table}[ht]
\centering
\small
\caption{Resnik similarity and co-occurrence similarity scores for different models (UK Biobank).}
\begin{tabular}{p{3cm}cc}
\toprule
\textbf{Model} & \makecell{\textbf{Resnik} \\ \textbf{Sim.}} & \makecell{\textbf{Co-occur.} \\ \textbf{Sim.}} \\
\Xhline{1pt}
\textbf{BioClinBERT} & 0.40 & 0.13 \\
\textbf{BioClinBERT$_{HA}$} & 0.50 & 0.15 \\
\textbf{BioGPT} & 0.48 & 0.25 \\
\textbf{BioGPT$_{HA}$} & 0.67 & 0.36 \\
\textbf{ClinBERT} & 0.40 & 0.18 \\
\textbf{ClinBERT$_{HA}$} & 0.51 & 0.21 \\
\textbf{Cui2Vec} & 0.37 & 0.55 \\
\textbf{GloVe} & 0.55 & 0.57 \\
\textbf{Node2Vec} & \textbf{0.70} & 0.46 \\
\textbf{GAT} & 0.43 & 0.31 \\
\Xhline{1pt}
\textbf{KEEP} & 0.68 & \textbf{0.62} \\
\bottomrule
\end{tabular}
\label{tab:similarity_scores}
{\footnotesize Model$_{HA}$ refers to Hierarchy Aware embeddings. Column names: "Sim." = Similarity, "Co-occur." = Co-occurrence}
\end{table}

\subsection{Intrinsic Evaluation}
Table \ref{tab:clinical_conditions_comparison} presents performance comparisons across eight clinical conditions. KEEP demonstrated superior performance in capturing clinically meaningful relationships, outperforming both language models and traditional embedding methods. Our method achieved the highest average rank of 1.19 across all evaluated conditions (p=0.01). The advantage of combining structural and empirical relationships is particularly evident in conditions like Type 1 diabetes mellitus, where our method achieved a performance score of 0.93, substantially outperforming both GloVe (0.82) and Node2Vec (0.27).

Language models demonstrated notably weaker performance, particularly for specific conditions such as Prematurity (scores ranging from 0.00-0.11) and Obesity (scores ranging from 0.00-0.20). The addition of hierarchical information to these models did not yield consistent improvements in performance.

\begin{table*}[htbp]
\centering
\small
\caption{\textbf{Intrinsic evaluation}: Accuracy of embedding models in identifying comorbid conditions across eight test diseases. Values represent the proportion of correctly identified comorbid conditions (UK Biobank).}
\setlength{\tabcolsep}{3pt}
\renewcommand{\arraystretch}{1.2}
\begin{tabular}{p{2.5cm}cccccccccc}
\toprule
\thead{} & \thead{Asthma} & \thead{Obesity} & \thead{Pre-\\maturity} & \thead{Renal\\Tx Reject.} & \thead{Schizo-\\phrenia} & \thead{Resp.\\Tumor} & \thead{T1DM} & \thead{T2DM} & \thead{Mean\\Rank} & \thead{p-val} \\ 
\Xhline{1pt}
\textbf{BioClinBERT} & 0.54 & 0.12 & 0.00 & 0.35 & 0.43 & 0.00 & 0.27 & 0.20 & 7.63 & 0.03 \\
\textbf{BioClinBERT$_{HA}$} & 0.12 & 0.02 & 0.14 & 0.00 & 0.40 & 0.31 & 0.02 & 0.08 & 9.75 & 0.01 \\ 
\textbf{ClinBERT} & 0.24 & 0.09 & 0.00 & \textbf{0.72} & 0.46 & 0.25 & 0.52 & 0.33 & 6.75 & 0.64 \\
\textbf{ClinBERT$_{HA}$} & 0.22 & 0.07 & 0.14 & 0.02 & 0.49 & 0.45 & 0.05 & 0.00 & 8.50 & 0.04 \\
\textbf{BioGPT} & 0.37 & 0.16 & 0.14 & 0.30 & 0.23 & 0.54 & 0.27 & 0.36 & 7.43 & 0.04 \\
\textbf{BioGPT$_{HA}$} & 0.39 & 0.22 & 0.57 & 0.43 & 0.43 & 0.66 & 0.18 & 0.16 & 6.13 & 0.87 \\
\textbf{Cui2vec} & 0.61 & 0.74 & \textbf{0.90} & 0.43 & 0.69 & \textbf{0.72} & 0.91 & 0.69 & 2.63 & 0.01 \\
\textbf{GloVe} & 0.54 & 0.64 & 0.76 & 0.70 & \textbf{0.86} & 0.68 & 0.82 & 0.69 & 2.63 & 0.01 \\
\textbf{Node2Vec} & 0.61 & 0.09 & 0.62 & 0.50 & 0.31 & 0.34 & 0.27 & 0.15 & 6.56 & 0.64 \\
\textbf{GAT} & 0.65 & 0.26 & 0.19 & 0.0 & 0.40 & 0.28 & 0.64 & 0.69 & 5.88 & 0.93 \\
\Xhline{1pt}
\textbf{KEEP} & \textbf{0.78} & \textbf{0.93} & 0.57 & 0.70 & 0.69 & 0.68 & \textbf{0.93} & \textbf{0.79} & \textbf{2.13} & 0.01 \\
\bottomrule
\end{tabular}
\label{tab:clinical_conditions_comparison}
{\footnotesize Model$_{\text{HA}}$ refers to Hierarchy Aware embeddings. Column names: "Tx" = Transplant, "Resp." = Respiratory, "T1DM" = Type 1 Diabetes Mellitus, "T2DM" = Type 2 Diabetes Mellitus, "p-val" = p-value.}
\end{table*}

\subsection{Extrinsic Evaluation}
Table \ref{tab:embedding_comparison} presents AUPRC results for the downstream clinical prediction tasks. KEEP demonstrated superior performance compared to baseline methods, achieving a mean rank of 1.62 across multiple clinical outcomes (p=0.02). Additional metrics reported in Tables \ref{tab:embedding_comparison_loss}-\ref{tab:embedding_comparison_f1} show that KEEP achieved the lowest test loss, second-highest AUC, highest MCC, and highest F1 score. All performance differences were statistically significant ($p<0.05$). 

Among the baseline approaches evaluated, traditional methods demonstrated consistency in their performance. Cui2Vec and node2vec achieved strong results across the prediction tasks. In contrast, LM performance showed greater variation. While BioGPT emerged as the second-strongest performer overall—likely attributable to its modern architecture and comprehensive training dataset—ClinBERT consistently delivered suboptimal results. Our attempts to enhance language model performance through hierarchical descriptions did not yield consistent improvements, suggesting that simple concatenation may not capture the complexity of clinical relationships in prediction tasks.
\begin{table*}[htpb]
\centering
\footnotesize
\caption{\textbf{Extrinsic evaluation}: AUPRC scores}
\vspace{2mm}
\setlength{\tabcolsep}{3pt}
\renewcommand{\arraystretch}{1.5}
\begin{tabular}{>{\bfseries}p{1.2cm}ccccccccc}
\toprule
\thead{} & \thead{All DM} & \thead{Periph.\\Vasc.} & \thead{Kidney} & \thead{Eye} & \thead{Neuro-\\pathy} & \thead{MI} & \thead{Renal\\Fail.} & \thead{Re-\\admit} & \thead{Mean \\ Rank} \\  
\Xhline{1pt}
\makecell[l]{BioClin\\BERT} & $0.59{\pm 0.003}$ & $0.21{\pm 0.013}$ & $0.29{\pm 0.002}$ & $0.45{\pm 0.002}$ & $0.25{\pm 0.002}$ & $0.20{\pm 0.003}$ & $0.35{\pm 0.002}$ & $0.073{\pm 0.005}$ & 5.88\\
\makecell[l]{BioClin\\BERT$_{H}$} & $0.55{\pm 0.002}$ & $0.14{\pm 0.016}$ & $0.30{\pm 0.006}$ & $0.44{\pm 0.001}$ & $0.24{\pm 0.001}$ & $0.20{\pm 0.001}$ & $0.34{\pm 0.001}$ & $0.062{\pm 0.011} $ & 7.81 \\
\makecell[l]{Clin\\BERT} & $0.51{\pm 0.010}$ & $0.10{\pm 0.011}$ & $0.28{\pm 0.010}$ & $0.40{\pm 0.010}$ & $0.20{\pm 0.013}$ & $0.20{\pm 0.005}$ & $0.18{\pm 0.025}$ & $0.045 {\pm 0.001}$& $10.19^*$ \\ 
\makecell[l]{Clin\\BERT$_{H}$} & $0.52{\pm 0.003}$ & $0.15{\pm 0.006}$ & $0.27{\pm 0.008}$ & $0.41{\pm 0.009}$ & $0.23{\pm 0.005}$ & $0.20{\pm 0.001}$ & $0.28{\pm 0.018}$ & $0.045 {\pm 0.002}$& $9.44^*$ \\
\makecell[l]{Bio\\GPT} & $0.60{\pm 0.001}$ & $0.20{\pm 0.006}$ & $0.33{\pm 0.002}$ & $\textbf{0.48}{\pm 0.001}$ & $0.22{\pm 0.001}$ & $0.21{\pm 0.000}$ & $\textbf{0.36}{\pm 0.001}$ & $0.079 {\pm 0.007}$& 4.12  \\
\makecell[l]{Bio\\GPT$_{H}$} & $\textbf{0.61}{\pm 0.001}$ & $0.18{\pm 0.005}$ & $0.32{\pm 0.002}$ & $0.41{\pm 0.006}$ & $0.26{\pm 0.005}$ & $\textbf{0.22}{\pm 0.001}$ & $0.31{\pm 0.019}$ & $0.077{\pm 0.017}$& 5.88 \\
\makecell[l]{Cui2vec} & $0.57{\pm 0.002}$ & $0.17{\pm 0.009}$ & $0.32{\pm 0.003}$ & $0.47{\pm 0.001}$ & $0.26{\pm 0.004}$ & $\textbf{0.22}{\pm 0.001}$ & $\textbf{0.36}{\pm 0.002}$ & $0.095 {\pm 0.004}$& $4.00^*$ \\
\makecell[l]{Node2\\Vec} & $0.60{\pm 0.001}$ & $0.20{\pm 0.010}$ & $0.34{\pm 0.002}$ & $0.46{\pm 0.001}$ & $0.31{\pm 0.002}$ & $\textbf{0.22}{\pm 0.001}$ & $0.34{\pm 0.001}$ & $0.103 {\pm 0.004}$& $4.00^*$ \\
\makecell[l]{GloVe} & $0.54{\pm 0.001}$ & $0.13{\pm 0.012}$ & $0.32{\pm 0.002}$ & $0.46{\pm 0.001}$ & $0.27{\pm 0.003}$ & $\textbf{0.22}{\pm 0.001}$ & $\textbf{0.36}{\pm 0.001}$ & $0.113 {\pm 0.003}$& 5.19 \\
\makecell[l]{GAT} & $0.57{\pm 0.003}$ & $0.22{\pm 0.011}$ & $0.26{\pm 0.013}$ & $0.47{\pm 0.004}$ & $0.18{\pm 0.006}$ & $0.18{\pm 0.002}$ & $0.33{\pm 0.005}$ & $0.113 {\pm 0.003}$& 7.88 \\
\Xhline{1pt}
\makecell[l]{KEEP} & $\textbf{0.61}{\pm 0.001}$ & $\textbf{0.28}{\pm 0.003}$ & $\textbf{0.36}{\pm 0.002}$ & $\textbf{0.48}{\pm 0.001}$ & $\textbf{0.32}{\pm 0.002}$ & $\textbf{0.22}{\pm 0.001}$ & $0.35{\pm 0.001}$ & $\textbf{0.122} {\pm 0.002}$& $\textbf{1.62}^*$\\
\bottomrule
\end{tabular}
\label{tab:embedding_comparison}
{\footnotesize Model$_{\text{H}}$ refers to Hierarchy Aware embeddings. Column names: "DM" = Diabetes Mellitus, "Periph. Vasc." = Peripheral Vascular Disease, "Neuro." = Neuropathy, "MI" = Myocardial Infarction, "Renal Fail." = Renal Failure, "Readmit" = 30d readmission (MIMIC-IV), * = p$<0.05$}
\end{table*}

\subsection{Runtime}
KEEP offers a significant advantage in computational efficiency. Generating embeddings for the entire disease terminology using data from 500,000 patients was completed in under 2 hours on a single NVIDIA L40S GPU with 46 GB of memory (see Section~\ref{app:model_params} for training details). In contrast, fine-tuning language models (LMs) often requires several hours and multiple GPUs, making KEEP a more practical and resource-efficient solution.

\section{Discussion}
Our results demonstrate two key findings about medical code embeddings. First, embeddings trained directly on institutional data can outperform LMs when fine-tuning is not feasible. This advantage likely stems from their ability to learn dataset-specific relationships. Second, our approach shows that carefully combining structural knowledge with empirical patterns through regularization creates more effective embeddings for real-world clinical prediction tasks. KEEP produces final embeddings without auxiliary or end-to-end training for specific tasks. This makes KEEP complementary to existing methods. For instance, GRAM \citep{choi2016multi} and HealthGAT \citep{piya2024healthgat} initialize embeddings with GloVe and node2vec, respectively, before conducting task-specific training. This suggests KEEP embeddings could serve as an initialization for such methods.

These findings have important implications for healthcare institutions. KEEP provides a practical approach for generating high-quality disease representations without requiring extensive computational resources or massive datasets. This efficiency makes our method particularly suitable for resource-constrained healthcare settings.

KEEP's flexibility is another key advantage. Through the regularization parameter $\lambda$, institutions can tune the contribution of knowledge graphs and empirical patterns based on their specific needs and data characteristics. Healthcare systems with limited patient data can emphasize knowledge graphs through a higher $\lambda$ value, while those with rich clinical data might benefit from a lower $\lambda$ to prioritize empirical patterns. By integrating both knowledge graph structures and local clinical patterns, institutions can create embeddings that better reflect their specific patient populations while maintaining alignment with established medical knowledge. This balance is particularly valuable for clinical prediction tasks and rare diseases, where understanding both theoretical disease relationships and real-world manifestations is crucial.

\subsection{Complementarity LM's}
Rather than competing with LM-based approaches, our approach serves as a complementary tool that addresses specific limitations of language models in handling structured medical data. Our framework bridges structured medical knowledge and unstructured data representations in two key ways. First, our embeddings can be integrated with LM-derived representations through multimodal integration techniques \citep{ebrahimi2023lanistr}. Second, our method could enhance LM performance by providing domain-aware initialization for fine-tuning, potentially improving convergence and reducing data requirements while enabling more effective incorporation of medical ontologies and relationships \citep{hewitt2021initializing, fatemi2023talk}.

\subsection{Limitations and Future Work}
Our knowledge graph implementation currently utilizes only hierarchical relationships between diseases, omitting other important clinical connections such as causal and associative relationships. Incorporating these additional relationship types could provide richer representations of disease interactions \citep{shen2019hpo2vec+}. Also, the current approach does not capture the temporal dynamics of disease progression. Future implementations could address this limitation by incorporating visit-level information rather than aggregating across entire patient histories. Finally, we focus only on disease codes which provides a limited view of patient health. Future work could explore methods for generating comprehensive patient representations that use multiple domains such as medications, labs, and observations.

\newpage
\bibliography{bibliography}
\appendix
\onecolumn

\section{Embedding Creation}
\begin{algorithm}[H]
\caption{KEEP}
\label{alg:KEEP_pseudocode}
\SetAlgoLined
\KwIn{Knowledge Graph $G = (V, E)$, EHR data $D$, Regularization parameter $\lambda$, Learning rate $\eta$, Number of epochs $T$}
\KwOut{Final embeddings $W = \{w_i\}$ for all $i \in V$}

\vspace{0.5em}
\textbf{Stage 1: Knowledge Graph Embedding}\;
\\
\vspace{0.3em}
Initialize node2vec parameters: \textit{window size, walk length, number of walks per node}\;
\vspace{0.3em}

\ForEach{node $v \in V$}{
    Generate random walks starting from $v$\;
}
\vspace{0.3em}
Train a skip-gram model using random walks to obtain node2vec embeddings:
\[
W^{\text{n2v}} = \{w_i^{\text{n2v}}\}
\]

\vspace{0.5em}
\textbf{Stage 2: EHR-Enhanced Representation Learning}\;
\\
\vspace{0.3em}
Construct co-occurrence matrix $X$ from EHR data $D$\;
\\
\vspace{0.3em}
Initialize GloVe embeddings with node2vec: 
\[
w_i \leftarrow w_i^{\text{n2v}}, \quad \forall i \in V
\]
\\
\vspace{0.5em}
\textbf{Training Loop:}\;
\\
\For{$t = 1$ to $T$}{
    Compute the \textbf{GloVe loss}:
    \[
    L_{\text{GloVe}} = \sum_{i,j=1}^{V} f(X_{ij}) \left( w_i^\top \tilde{w}_j + b_i + \tilde{b}_j - \log X_{ij} \right)^2
    \]

    Compute the \textbf{Regularization loss}:
    \[
    L_{\text{reg}} = \lambda \sum_{i=1}^{V} ||w_i - w_i^{\text{n2v}}||^2
    \]

    Compute the total objective:
    \[
    J(W) = L_{\text{GloVe}} + L_{\text{reg}}
    \]

    Update embeddings using AdamW:
    \[
    w_i \leftarrow w_i - \eta \cdot \nabla J(W), \quad \forall i \in V
    \]
}

\vspace{0.3em}

\KwRet{Final learned embeddings $W = \{w_i\}$}
\end{algorithm}

\subsection{Node2Vec}
\label{app:n2v}
Node2vec embeddings were generated using random walks on the OMOP knowledge graph.
\subsubsection{Graph Creation}
\label{app:graph}
We construct the graph by first filtering for disease-related concepts. Disease concepts are identified using the "CONCEPT" table, selecting entries with "domain\_name" "Condition". To limit graph complexity, we filter concepts based on their hierarchical distance from the root node, "Disease" (concept ID: 4274025). Using the "CONCEPT\_ANCESTOR" table, we calculate the minimum number of hierarchical levels separating each disease concept from the root node. Concepts with more than five levels of separation are excluded from the graph.

\subsection{Model parameters}
\label{app:model_params}
We used the following parameters to train Node2Vec:
\begin{table}[ht]
\centering
\small
\caption{Hyperparameters for Node2Vec Training}
\begin{tabular}{ll}
\toprule
\textbf{Hyperparameter} & \textbf{Value} \\ \Xhline{1pt}
Embedding dimension     & 100            \\
Walk length             & 30             \\
Number of walks         & 750            \\
p, q       & 1  \\
Window size             & 10             \\
Minimum count           & 1              \\
Batch size              & 4096           \\
\bottomrule
\end{tabular}
\label{tab:embedding_hyperparameters}
\end{table}
\subsection{GloVe}
\label{app:glove}
GloVe embeddings were generated using co-occurrence data from UKBB in OMOP format.
\subsection{Co-occurrence matrix}
\label{app:cooc_mat}
We construct the co-occurrence matrix using the same codes from our previous graph analysis. Instead of excluding codes more than five levels away from the root node, we implement a roll-up procedure that maps each code to its parent codes present in the graph. Adopting a dense roll-up approach, we map every code to \emph{all} of its parents, creating multiple entries when a code has multiple parent nodes. Using this dataset we create a co-occurrence matrix $X$ where each entry $X_{ij}$ represents the frequency of diseases $i$ and $j$ occurring together in a patient's records. To establish the presence of a disease, we require at least two occurrences in a patient’s history. Co-occurrence is determined based on the patient’s complete medical history, rather than being restricted to individual visits.

\subsubsection{Model parameters}
We used the following parameters to train GloVe:
\begin{table}[ht]
\centering
\small
\caption{Hyperparameters for GloVe Training}
\begin{tabular}{ll}
\toprule
\textbf{Hyperparameter} & \textbf{Value} \\ \Xhline{1pt}
Embedding dimension     & 100  \\
Learning rate           & 0.05                          \\
Number of epochs        & 300                           \\
Batch size              & 1024                          \\
X$_{max}$                  & 75$^{th}$ percentile \\
$\alpha$ & 0.75      \\
$\lambda$ & $1 \times 10^{-3}$ \\
\bottomrule
\end{tabular}
\label{tab:training_hyperparameters}
\end{table}

Note the regularization parameter $\lambda$ is only used in KEEP.

\subsection{Pre-trained embeddings}
\label{app:pre_trained_emb}
We utilized three pre-trained biomedical language models to extract disease embeddings in two ways: \textbf{Basic} and \textbf{Hierarchy-aware} description embeddings. The basic version uses only the description for the disease code. To incorporate hierarchical context, we constructed ancestry paths for each disease concept by combining descriptions from its hierarchy, specifically parent and grandparent concepts. The format for the hierarchical path was:
\begin{quote}
    \texttt{"[Disease]; is a [Parent]; is a [Grandparent]"}
\end{quote}

As we maintain the embedding dimensions for the BERT and GPT-based models, they have the following dimensions:
\begin{table}[h!]
\small
\caption{Embedding sizes for Pre-trained models.}
\centering
\begin{tabular}{lc}
\toprule
\textbf{Model Name}      & \textbf{Embedding Size} \\ \Xhline{1pt}
BioClinicalBERT        & 768                     \\ 
ClinicalBERT             & 768                     \\ 
BioGPT                   & 1024                    \\ 
\bottomrule
\end{tabular}
\label{tab:model_embeddings_dim}
\end{table}

To mitigate the risks overfitting and ensure numerical stability, we applied L2 normalization to these embeddings, following the recommendations of \citep{openai_embeddings_faq} and \citep{loshchilov2024ngpt}.

\section{Experiment setup}
\subsection{Assessing the Impact of Regularization}
\label{app:impact_reg}
For each code, we identified the ten most similar concepts based on cosine similarity, along with 150 randomly sampled concepts. For each selected concept, we computed its Resnik similarity (refer to Section \ref{background_omop} for calculation details) and its co-occurrence frequency with the original code. We the measured the correlation between cosine similarity and both Resnik similarity and co-occurrence values. To ensure statistical robustness, we  repeated the experiment 250 times report the median correlation values across all runs.

\subsection{Intrinsic evaluation}
\label{app:int_eval}
To evalute the quality of the embedding space, we assessed the embeddings' ability to identify known medical relationships, adapting the approach described by \citep{beam2020clinical}. This benchmark evaluates how well the embeddings capture established medical knowledge and real-world disease associations by measuring their ability to discriminate between known disease relationships.

For each core concept, we constructed positive pairs that included known relationships, such as complications, comorbidities, child diseases, and synonyms. Negative pairs were generated by randomly sampling concept pairs from the terminology, ensuring no overlap with the positive set. Discrimination performance was quantified as the proportion of known relationships achieving higher cosine similarity than 95\% of random pairs for each core concept. Formally, this can be represented as $P\left(\text{sim}(+{\text{pair}}) > \text{sim}(-{\text{pair}}) \right)$ where $\text{sim}(\cdot)$ denotes the cosine similarity between embeddings. To ensure robust estimates, we employed repeated the experiment times and used bootstrap sampling (1000 iteration) to calculate median scores, confidence intervals, and statistical significance. Table \ref{tab:core_concepts_comorbidities} shows the OMOP concept IDs for the core concept and the known relationship concepts used in the intrinsic evaluation.

\begin{table*}
\centering
\phantomsection
\caption{\textbf{Intrinsic Evaluation}: OMOP concept IDs for core concepts along with their associated synonyms, child diseases, complications, and comorbidities used for intrinsic evaluation.}
\label{tab:core_concepts_comorbidities}
\scriptsize
\begin{tabular}{|>{\raggedright\arraybackslash}p{1.3cm}|>{\raggedright\arraybackslash}p{1.3cm}|>{\raggedright\arraybackslash}p{3.2cm}|>{\raggedright\arraybackslash}p{4.2cm}|>{\raggedright\arraybackslash}p{4.5cm}|}

\hline
\textbf{Core Concept} & \textbf{Synonyms} & \textbf{Child Diseases} & \textbf{Complications} & \textbf{Comorbidities} \\ \hline
Asthma (317009) & $\cdot$ & 
4123253, 4051466, 4110051, 4233784, 40483397, 4145497, 312950, 443801, 4191479, 4155469, 4155470, 4119298, 4155468 & 
257581, 45769438 & 
4148368, 260123, 255573, 4214438, 4280726, 255841, 4182370, 4335888, 4112367, 4305500, 4283893, 256439, 256448, 259848, 4110489, 4110492, 4329087, 4223595, 380111, 4138403, 4195007, 257007, 133835, 257012, 42537251, 42537252 \\ \hline
Obesity (433736) & $\cdot$ & 
434005, 4189665, 4217557, 4087487 & 
4059290, 4026131, 436940, 4100857, 40482277, 40484532 & 
4079750, 314378, 4291025, 321318, 4083696, 313459, 373503, 201820, 44809569, 381591, 321588, 315286, 4079749, 314666, 4112022, 442604, 44809026, 4209293, 73553, 374384, 4110196, 320128, 442588, 4147779, 4098302, 381316, 4045734, 443454, 4149320, 4186397, 439150, 435524, 443732, 201826, 4185932, 319844, 312327, 80180, 316866, 4043734, 432867, 319034, 4314692, 40481943, 46271022, 4170226, 4151170, 317576 \\ \hline
Premature rupture of membranes (194702) & $\cdot$ & 
4064296, 45772076, 4060691 & 
$\cdot$ & 
36712695, 434111, 4058243, 43530976, 4049790, 438815, 435656, 4042220, 437623, 72693, 435875, 4146482, 433823, 4118910, 443247, 4304781, 4024659, 4062791 \\ \hline
Renal transplant rejection (4128369) & 
4309006, 4324887, 4309320, 4127554 & 
$\cdot$ & 
201461, 140362, 4324887, 36715574 & 
443612, 443614, 4242411, 443731, 4030664, 4056478, 4264718, 198185, 4059452, 4153654, 443597, 443601, 192359, 437247, 4126305, 312358, 4206115, 4118795, 4128221, 196455, 606956, 200687, 4030518, 192964, 201313, 4269363, 45757752, 4131748, 443919, 35623051, 252365, 197320, 45757772, 4220238, 193253, 42536547, 435308, 195556, 443611, 35624383, 46271022, 192279 \\ \hline
Schizophrenia (435783) & 
4100365 & 
4286201, 435219, 4100365, 4153292 & 
4286201, 4101149, 432590, 4100247, 4102670, 433450, 618641, 436073, 4335169 & 
4152280, 4159691, 434613, 4308866, 4004672, 438028, 434223, 440383, 4098302, 40481346, 442077, 4149320, 4103574, 4149321, 4282316, 36713290, 4314692, 435520, 434819, 4239381, 4338031, 4151170 \\ \hline
Tumor of respiratory system (40491439) & $\cdot$ & 
4180795, 45769033, 4054511, 4112735, 4093957, 4242982, 78093, 4247331, 4128888, 4113116, 252840, 4114353, 4240153, 40493428, 4055270, 261528, 4054836 & 
192568, 4129382, 434875, 4317284, 319049, 317003, 4246451, 441258, 4114341, 72266, 320342, 4256228, 443588, 4131304, 439751, 4102360, 4180795, 432851, 4233244, 318096, 4130839 & 
4124677, 257011, 255573, 37206139, 4307774, 44807895, 255841, 4317284, 4112341, 4170143, 4000938, 4115044, 4119786, 4110056, 4208807, 4195694, 4110479, 253506, 4131304, 40491473, 132797, 4175297, 260139, 4112357, 4148204, 256451, 257004 \\ \hline
Type 1 diabetes mellitus (201254) & 
4130161, 201820, 4034959, 44808373, 4308509, 4008576, 4311629 & 
$\cdot$ & 
42536605, 443727, 4058243, 4029423, 443735, 376979, 4174262, 4030664, 42538169, 377821, 380097, 4082346, 4034959, 4206115, 443730, 4128221, 318712, 200687, 4016047, 321822, 435216, 442793, 4044391, 376112, 24609, 4214376, 4048028, 4209145, 4174977, 30361, 443767, 192279 & 
315286, 4185932, 319844, 376686, 443612, 4217557, 443614, 433736, 443597, 443601, 443611, 46271022 \\ \hline
Type 2 diabetes mellitus (201826) & 
44808385, 201820, 4034959, 44808373, 4308509, 4008576, 4311629 & 
$\cdot$ & 
376979, 443731, 443729, 380097, 376065, 4082346, 4206115, 443730, 321822, 442793, 4044391, 376112, 443733, 443732, 4174977, 4223739, 443767, 192279, 4221487 & 
434005, 319826, 443612, 443614, 373503, 321318, 4119612, 44809569, 4087487, 4200991, 321588, 315286, 314666, 4217557, 442604, 44809026, 4209293, 443597, 443601, 374384, 374034, 320128, 4098302, 4149320, 4186397, 43021237, 4185932, 319844, 4149321, 312327, 316866, 433736, 4043734, 432867, 4124836, 3655355, 4314692, 443611, 46271022, 4170226, 4151170, 317576 \\ \hline
\end{tabular}
\scriptsize{All child concepts of codes shown are also used}
\end{table*}

\newpage
\subsection{Extrinsic evaluation}
\label{app:ext_eval}
Below we detail the evaluation methodology. Implementation code is available at \url{https://github.com/G2Lab/keep}
\subsubsection{Data Preprocessing and Cohort Creation}
We required disease concepts to appear a minimum of two times in a patient's history for inclusion. Table \ref{tab:outcome_codes} shows the OMOP concept IDs used to define the cohort and outcome. For patients with positive outcomes, we required two instances of the outcome code and censored all data after the first occurrence to ensure only preceding medical history was considered. Sequence lengths were limited to a maximum of 20 codes, with random sampling applied to the small subset of sequences exceeding this limit. The final dataset was partitioned using label stratification to maintain outcome distributions: an initial 80-20 train-test split was performed, followed by an 80-20 subdivision of the training data, yielding final proportions of 64\% training, 16\% validation, and 20\% testing data. 

 \begin{table}[htbp]
\centering
\small
\caption{\textbf{Extrinsic evaluation}: OMOP concept IDs used to define cohort, outcome and exclusions}
 \label{tab:outcome_codes}
\begin{tabular}{>{\raggedright\arraybackslash}p{1.3cm}>{\raggedright\arraybackslash}p{3.8cm}>{\raggedright\arraybackslash}p{1.6cm}}
\toprule
\textbf{Inclusion}& \textbf{Outcome} & \textbf{Exclusion} \\ \Xhline{1pt}
 201826 & 442793 & 201254\\
 201826 & 321822 & 201254\\
 201826 & 1992279 & 201254\\
 201826 & 443767 & 201254\\
201826 &  443730 & 201254\\
 321318 & 4329847 & 314666*\\
46271022 & 197320, 432961, 444044 & $\cdot$ \\
\bottomrule
\end{tabular}
\end{table}

\subsubsection{Model Architecture}
We implemented a transformer model with one encoder layer and four attention heads. Input dimensions match those of the corresponding embeddings. Regularization was applied through dropout (p=0.2). We employed attention-based pooling for sequence representation, followed by a binary classification layer.

\subsubsection{Training Protocol}
Models were trained with a batch size of 32 using AdamW optimization (weight decay=0.01) and cross-entropy loss. Learning rates were evaluated across nine values ranging from 1$e^{-5}$ to 1$e^{-1}$ in logarithmic increments using grid search. We trained for a maximum of 500 epochs with early stopping triggered after 5 epochs without validation loss improvement. We repeated each experiment 5 times and selected the learning rate and epoch that produced the lowest median validation loss. Final evaluation comprised 100 independent runs, with confidence intervals computed via bootstrap resampling (1000 iterations).

\subsubsection{Evaluation Metrics}
Model performance was primarily assessed using AUPRC AUC, MCC, F1 score, and Test loss. For statistical comparison between embedding methods, we employed the Wilcoxon signed-rank test on the performance ranks across all tasks.
\newpage

\section{Experiment results}
\label{app:exp_results}
Tables \ref{tab:embedding_comparison_loss}-\ref{tab:embedding_comparison_f1} show the results for Test loss, AUC, MCC, and F1 score in the extrinsic evaluation tasks.
\begin{table*}[htbp]
\centering
\footnotesize
\caption{\textbf{Extrinsic evaluation}: Test loss (lowest is best)}
\setlength{\tabcolsep}{3pt}
\renewcommand{\arraystretch}{1.5}
\begin{tabular}{>{\bfseries}p{1.4cm}ccccccccc}
\toprule
\thead{} & \thead{All DM} & \thead{Periph.\\Vasc.} & \thead{Kidney} & \thead{Eye} & \thead{Neuro-\\pathy} & \thead{MI} & \thead{Renal\\Fail.} & \thead{Re-\\admit} & \thead{Mean\\Rank} \\
\Xhline{1pt}
\makecell[l]{BioClin\\BERT} & $0.53{\pm 0.003}$ & $0.63{\pm 0.008}$ & $0.59{\pm 0.001}$ & $0.55{\pm 0.001}$ & $0.61{\pm 0.001}$ & $0.61{\pm 0.003}$ & $0.38{\pm 0.008}$ & $0.67{\pm 0.002}$ & 6.12 \\
\makecell[l]{BioClin\\BERT$_{H}$} & $0.55{\pm 0.001}$ & $0.65{\pm 0.007}$ & $0.57{\pm 0.003}$ & $0.55{\pm 0.001}$ & $0.62{ \pm 0.001}$ & $0.62{ \pm 0.002}$ & $0.37{ \pm 0.002}$ & $0.70{ \pm 0.003}$ & 7.75 \\
\makecell[l]{Clin\\BERT} & $0.57{ \pm 0.005}$ & $0.63{ \pm 0.008}$ & $0.57{ \pm 0.006}$ & $0.57{ \pm 0.004}$ & $0.62{ \pm 0.006}$ & $0.62{ \pm 0.005}$ & $0.56{ \pm 0.029}$ & $0.68{ \pm 0.001}$ & 9.12 \\ 
\makecell[l]{Clin\\BERT$_{H}$} & $0.57{ \pm 0.001}$ & $0.63{ \pm 0.004}$ & $0.59{ \pm 0.004}$ & $0.57{ \pm 0.005}$ & $0.61{ \pm 0.004}$ & $0.62{ \pm 0.001}$ & $0.43{ \pm 0.024}$ & $0.68{ \pm 0.001}$ & 8.50 \\
\makecell[l]{Bio\\GPT} & $0.52{ \pm 0.001}$ & $0.65{ \pm 0.007}$ & $0.56{ \pm 0.001}$ & $0.53{ \pm 0.001}$ & $0.59{ \pm 0.001}$ & $0.60{ \pm 0.000}$ & $0.37{ \pm 0.001}$ & $0.68{ \pm 0.010}$ & 4.38 \\
\makecell[l]{Bio\\GPT$_{H}$} & $0.52{ \pm 0.001}$ & $0.64{ \pm 0.006}$ & $0.56{ \pm 0.001}$ & $0.55{ \pm 0.002}$ & $0.58{ \pm 0.003}$ & $0.61{ \pm 0.002}$ & $0.44{ \pm 0.023}$ & $0.70{ \pm 0.008}$ & 6.06 \\
\makecell[l]{Cui2vec} & $0.54{ \pm 0.001}$ & $0.60{ \pm 0.006}$ & $0.60{ \pm 0.005}$ & $0.53{ \pm 0.001}$ & $0.59{ \pm 0.001}$ & $0.59{ \pm 0.000}$ & $0.38{ \pm 0.002}$ & $0.67{ \pm 0.002}$ & 4.31 \\
\makecell[l]{GloVe} & $0.55{ \pm 0.001}$ & $0.62{ \pm 0.008}$ & $0.56{ \pm 0.002}$ & $0.56{ \pm 0.002}$ & $0.58{ \pm 0.001}$ & $0.59{ \pm 0.000}$ & $0.37{ \pm 0.001}$ & $0.67{ \pm 0.003}$ & 4.56 \\
\makecell[l]{Node2\\Vec} & $0.53{ \pm 0.001}$ & $0.60{ \pm 0.005}$ & $0.58{ \pm 0.002}$ & $0.55{ \pm 0.001}$ & $0.58{ \pm 0.001}$ & $0.60{ \pm 0.001}$ & $0.38{ \pm 0.001}$ & $0.67{ \pm 0.002}$ & 4.56 \\
\makecell[l]{GAT} & $0.55{ \pm 0.002}$ & $0.60{ \pm 0.008}$ & $0.60{ \pm 0.006}$ & $0.55{ \pm 0.005}$ & $0.64{ \pm 0.020}$ & $0.62{ \pm 0.001}$ & $0.39{ \pm 0.005}$ & $0.69{ \pm 0.006}$ & 7.88 \\
\Xhline{1pt}
\makecell[l]{KEEP} & $0.52{ \pm 0.001}$ & $0.60{ \pm 0.002}$ & $0.57{ \pm 0.002}$ & $0.54{ \pm 0.001}$ & $0.57{ \pm 0.001}$ & $0.59{ \pm 0.000}$ & $0.38{ \pm 0.001}$ & $0.65{ \pm 0.001}$ & \textbf{2.75} \\
\bottomrule
\end{tabular}
\label{tab:embedding_comparison_loss}
\end{table*}

\begin{table*}[htbp]
\centering
\footnotesize
\caption{\textbf{Extrinsic Evaluation}: AUC scores}
\setlength{\tabcolsep}{3pt}
\renewcommand{\arraystretch}{1.5}
\begin{tabular}{>{\bfseries}p{1.4cm}ccccccccc}
\toprule
\thead{} & \thead{All DM} & \thead{Periph.\\Vasc.} & \thead{Kidney} & \thead{Eye} & \thead{Neuro-\\pathy} & \thead{MI} & \thead{Renal\\Fail.} & \thead{Re-\\admit} & \thead{Mean\\Rank} \\
\Xhline{1pt}
\makecell[l]{BioClin\\BERT} & $0.79{ \pm 0.001}$ & $0.70{ \pm 0.009}$ & $0.73{ \pm 0.001}$ & $0.78{ \pm 0.001}$ & $0.71{ \pm 0.001}$ & $0.74{ \pm 0.002}$ & $0.90{ \pm 0.002}$&  $0.65{ \pm 0.003}$ & 6.31  \\
\makecell[l]{BioClin\\BERT$_{H}$} & $0.77{ \pm 0.001}$ & $0.60{ \pm 0.016}$ & $0.75{ \pm 0.002}$ & $0.78{ \pm 0.001}$ & $0.70{ \pm 0.001}$ & $0.74{ \pm 0.001}$ & $0.90{ \pm 0.000}$&  $0.62{ \pm 0.026}$  & 7.25 \\
\makecell[l]{Clin\\BERT} & $0.75{ \pm 0.005}$ & $0.63{ \pm 0.018}$ & $0.73{ \pm 0.005}$ & $0.76{ \pm 0.004}$ & $0.69{ \pm 0.006}$ & $0.73{ \pm 0.007}$ & $0.67{ \pm 0.040}$& $0.61{ \pm 0.001}$  & 10.0  \\ 
\makecell[l]{Clin\\BERT$_{H}$} & $0.76{ \pm 0.001}$ & $0.72{ \pm 0.007}$ & $0.73{ \pm 0.005}$ & $0.76{ \pm 0.004}$ & $0.70{ \pm 0.005}$ & $0.73{ \pm 0.001}$ & $0.83{ \pm 0.029}$ &  $0.61{ \pm 0.001}$& 8.94  \\
\makecell[l]{Bio\\GPT} & $0.80{ \pm 0.000}$ & $0.72{ \pm 0.003}$ & $0.75{ \pm 0.001}$ & $0.79{ \pm 0.001}$ & $0.73{ \pm 0.001}$ & $0.75{ \pm 0.000}$ & $0.90{ \pm 0.000}$ & $0.67{ \pm 0.009}$ & \textbf{2.69}\\
\makecell[l]{Bio\\GPT$_{H}$} & \textbf{0.81}${ \pm 0.000}$ & \textbf{0.75}${ \pm 0.002}$ & \textbf{0.75}${ \pm 0.001}$ & $0.78{ \pm 0.001}$ & $0.73{ \pm 0.003}$ & $0.75{ \pm 0.001}$ & $0.85{ \pm 0.024}$ & $0.63{ \pm 0.032}$ & 4.00 \\
\makecell[l]{Cui2vec} & $0.78{ \pm 0.001}$ & $0.71{ \pm 0.012}$ & $0.74{ \pm 0.003}$ & \textbf{0.79}${ \pm 0.001}$ & $0.71{ \pm 0.002}$ & \textbf{0.76}${ \pm 0.000}$ & $0.90{ \pm 0.000}$ & $0.66{ \pm 0.002}$ & 3.69 \\
\makecell[l]{Node2\\Vec} & $0.79{ \pm 0.001}$ & $0.69{ \pm 0.008}$ & $0.74{ \pm 0.002}$ & $0.77{ \pm 0.001}$ & $0.71{ \pm 0.002}$ & $0.75{ \pm 0.001}$ & $0.90{ \pm 0.000}$ &  $0.66{ \pm 0.003}$& 6.44 \\
\makecell[l]{GloVe} & $0.77{ \pm 0.001}$ & $0.69{ \pm 0.018}$ & $0.74{ \pm 0.002}$ & $0.78{ \pm 0.001}$ & $0.71{ \pm 0.002}$ & $0.75{ \pm 0.000}$ & \textbf{0.90}${ \pm 0.000}$&   $0.66{ \pm 0.003}$& 4.50 \\
\makecell[l]{GAT} & $0.77{ \pm 0.001}$ & $0.69{ \pm 0.018}$ & $0.74{ \pm 0.002}$ & $0.78{ \pm 0.001}$ & $0.71{ \pm 0.002}$ & $0.75{ \pm 0.000}$ & \textbf{0.90}${ \pm 0.000}$&   $0.58{ \pm 0.029}$& 9.19 \\
\Xhline{1pt}
\makecell[l]{KEEP} & $0.80{ \pm 0.001}$ & $0.74{ \pm 0.002}$ & $0.74{ \pm 0.001}$ & $0.79{ \pm 0.001}$ & \textbf{0.74}${ \pm 0.002}$ & $0.75{ \pm 0.000}$ & $0.90{ \pm 0.000}$&   $\textbf{0.67}{ \pm 0.001}$& 3.00\\
\bottomrule
\end{tabular}
\label{tab:embedding_comparison_auc}
\end{table*}
\begin{table}[htbp]
\centering
\footnotesize
\caption{\textbf{Extrinsic Evaluation}: Mathews Correlation Coefficient}
\vspace{2mm}
\setlength{\tabcolsep}{3pt}
\renewcommand{\arraystretch}{1.5}
\begin{tabular}{>{\bfseries}p{1.6cm}ccccccccc}
\toprule
\thead{} & \thead{All DM} & \thead{Periph.\\Vasc.} & \thead{Kidney} & \thead{Eye} & \thead{Neuro-\\pathy} & \thead{MI} & \thead{Renal\\Fail.} & \thead{Re-\\admit} & \thead{Mean\\Rank} \\
\Xhline{1pt}
\makecell[l]{BioClin\\BERT} & $0.39{ \pm 0.009}$ & $0.19{ \pm 0.009}$ & $0.25{ \pm 0.002}$ & $0.29{ \pm 0.004}$ & $0.11{ \pm 0.001}$ & $0.20{ \pm 0.007}$ & $0.31{ \pm 0.008}$ & $0.09{ \pm 0.005}$ & 5.69  \\
\makecell[l]{BioClin\\BERT$_{HA}$} & $0.37{ \pm 0.004}$ & $0.14{ \pm 0.021}$ & $0.18{ \pm 0.005}$ & $0.30{ \pm 0.004}$ & $0.08{ \pm 0.001}$ & $0.21{ \pm 0.001}$ & $0.32{ \pm 0.001}$ & $0.00{ \pm 0.000}$ & 7.38 \\
\makecell[l]{Clin\\BERT} & $0.33{ \pm 0.013}$ & $0.13{ \pm 0.020}$ & $0.17{ \pm 0.009}$ & $0.24{ \pm 0.007}$ & $0.14{ \pm 0.010}$ & $0.18{ \pm 0.009}$ & $0.14{ \pm 0.031}$ & $0.06{ \pm 0.006}$ & 9.75  \\ 
\makecell[l]{Clin\\BERT$_{HA}$} & $0.35{ \pm 0.005}$ & $0.17{ \pm 0.011}$ & $0.25{ \pm 0.011}$ & $0.29{ \pm 0.008}$ & $0.24{ \pm 0.010}$ & $0.19{ \pm 0.002}$ & $0.26{ \pm 0.023}$ & $0.07{ \pm 0.000}$ & 6.38  \\
\makecell[l]{BioGPT} & $0.44{ \pm 0.003}$ & $0.15{ \pm 0.003}$ & $0.17{ \pm 0.002}$ & $0.32{ \pm 0.004}$ & $0.15{ \pm 0.002}$ & $0.21{ \pm 0.001}$ & $0.33{ \pm 0.001}$ & $0.08{ \pm 0.015}$ & 5.25 \\
\makecell[l]{BioGPT$_{HA}$} & $0.47{ \pm 0.004}$ & $0.19{ \pm 0.003}$ & $0.17{ \pm 0.002}$ & $0.27{ \pm 0.004}$ & $0.14{ \pm 0.003}$ & $0.19{ \pm 0.004}$ & $0.27{ \pm 0.024}$ & $0.03{ \pm 0.032}$ & 7.31 \\
\makecell[l]{Cui2vec} & $0.43{ \pm 0.005}$ & $0.13{ \pm 0.007}$ & $0.22{ \pm 0.004}$ & $0.29{ \pm 0.003}$ & $0.17{ \pm 0.002}$ & $0.22{ \pm 0.001}$ & $0.33{ \pm 0.002}$ & $0.09{ \pm 0.002}$ & 4.75  \\
\makecell[l]{GloVe} & $0.48{ \pm 0.003}$ & $0.13{ \pm 0.008}$ & $0.21{ \pm 0.003}$ & $0.35{ \pm 0.002}$ & $0.21{ \pm 0.004}$ & $0.20{ \pm 0.001}$ & $0.31{ \pm 0.001}$ & $0.11{ \pm 0.004}$ & 3.94 \\
\makecell[l]{Node2Vec} & $0.39{ \pm 0.005}$ & $0.13{ \pm 0.013}$ & $0.21{ \pm 0.003}$ & $0.27{ \pm 0.005}$ & $0.19{ \pm 0.002}$ & $0.22{ \pm 0.001}$ & $0.34{ \pm 0.001}$ & $0.11{ \pm 0.004}$ & 4.94 \\
\makecell[l]{GAT} & $0.36{ \pm 0.008}$ & $0.12{ \pm 0.013}$ & $0.17{ \pm 0.009}$ & $0.33{ \pm 0.011}$ & $0.20{ \pm 0.084}$ & $0.17{ \pm 0.005}$ & $0.32{ \pm 0.004}$ & $0.03{ \pm 0.023}$ & 7.75 \\
\Xhline{1pt}
\makecell[l]{KEEP} & $0.41{ \pm 0.002}$ & $0.20{ \pm 0.003}$ & $0.26{ \pm 0.004}$ & $0.35{ \pm 0.003}$ & $0.18{ \pm 0.003}$ & $0.21{ \pm 0.001}$ & $0.33{ \pm 0.001}$ & $0.11{ \pm 0.004}$ & \textbf{2.88} \\
\bottomrule
\end{tabular}
\label{tab:embedding_comparison_mcc}
\end{table}

\begin{table*}[htbp]
\centering
\footnotesize
\caption{\textbf{Extrinsic Evaluation}: F1 score}
\setlength{\tabcolsep}{3pt}
\renewcommand{\arraystretch}{1.5}
\begin{tabular}{>{\bfseries}p{1.4cm}ccccccccc}
\toprule
\thead{} & \thead{All DM} & \thead{Periph.\\Vasc.} & \thead{Kidney} & \thead{Eye} & \thead{Neuro-\\pathy} & \thead{MI} & \thead{Renal\\Fail.} & \thead{Re-\\admit} & \thead{Mean\\Rank} \\
\Xhline{1pt}
\makecell[l]{BioClin\\BERT} & $0.69{ \pm 0.010}$ & $0.58{ \pm 0.005}$ & $0.62{ \pm 0.001}$ & $0.63{ \pm 0.004}$ & $0.47{ \pm 0.004}$ & $0.55{ \pm 0.005}$ & $0.57{ \pm 0.003}$ & $0.48{ \pm 0.008}$ & 5.12  \\
\makecell[l]{BioClin\\BERT$_{H}$} & $0.68{ \pm 0.003}$ & $0.56{ \pm 0.010}$ & $0.55{ \pm 0.003}$ & $0.64{ \pm 0.004}$ & $0.42{ \pm 0.003}$ & $0.58{ \pm 0.003}$ & $0.56{ \pm 0.002}$ & $0.49{ \pm 0.000}$ & 6.38 \\
\makecell[l]{Clin\\BERT} & $0.65{ \pm 0.011}$ & $0.56{ \pm 0.009}$ & $0.55{ \pm 0.005}$ & $0.59{ \pm 0.007}$ & $0.56{ \pm 0.005}$ & $0.50{ \pm 0.018}$ & $0.53{ \pm 0.011}$ & $0.40{ \pm 0.017}$ & 9.00  \\ 
\makecell[l]{Clin\\BERT$_{H}$} & $0.67{ \pm 0.003}$ & $0.57{ \pm 0.005}$ & $0.62{ \pm 0.006}$ & $0.63{ \pm 0.006}$ & $0.62{ \pm 0.005}$ & $0.52{ \pm 0.002}$ & $0.56{ \pm 0.008}$ & $0.38{ \pm 0.000}$ & 6.19  \\
\makecell[l]{Bio\\GPT} & $0.72{ \pm 0.002}$ & $0.54{ \pm 0.003}$ & $0.54{ \pm 0.002}$ & $0.65{ \pm 0.003}$ & $0.54{ \pm 0.002}$ & $0.54{ \pm 0.002}$ & $0.57{ \pm 0.003}$ & $0.38{ \pm 0.060}$ & 6.75 \\
\makecell[l]{Bio\\GPT$_{H}$} & $0.73{ \pm 0.002}$ & $0.57{ \pm 0.002}$ & $0.55{ \pm 0.003}$ & $0.61{ \pm 0.003}$ & $0.50{ \pm 0.004}$ & $0.48{ \pm 0.008}$ & $0.57{ \pm 0.012}$ & $0.50{ \pm 0.009}$ & 6.25 \\
\makecell[l]{Cui2vec} & $0.72{ \pm 0.002}$ & $0.52{ \pm 0.004}$ & $0.60{ \pm 0.003}$ & $0.62{ \pm 0.003}$ & $0.56{ \pm 0.002}$ & $0.54{ \pm 0.001}$ & $0.60{ \pm 0.003}$ & $0.46{ \pm 0.004}$ & 5.62  \\
\makecell[l]{GloVe} & $0.69{ \pm 0.002}$ & $0.54{ \pm 0.007}$ & $0.59{ \pm 0.002}$ & $0.60{ \pm 0.006}$ & $0.58{ \pm 0.002}$ & $0.54{ \pm 0.001}$ & $0.59{ \pm 0.002}$ & $0.50{ \pm 0.004}$ & 5.25 \\
\makecell[l]{Node2\\Vec} & $0.73{ \pm 0.001}$ & $0.53{ \pm 0.006}$ & $0.59{ \pm 0.002}$ & $0.67{ \pm 0.001}$ & $0.59{ \pm 0.002}$ & $0.51{ \pm 0.002}$ & $0.57{ \pm 0.001}$ & $0.49{ \pm 0.004}$ & 4.88 \\
\makecell[l]{GAT} & $0.68{ \pm 0.006}$ & $0.52{ \pm 0.012}$ & $0.57{ \pm 0.008}$ & $0.66{ \pm 0.009}$ & $0.59{ \pm 0.044}$ & $0.47{ \pm 0.009}$ & $0.57{ \pm 0.003}$ & $0.47{ \pm 0.021}$ & 6.88 \\
\Xhline{1pt}
\makecell[l]{KEEP} & $0.70{ \pm 0.001}$ & $0.59{ \pm 0.002}$ & $0.62{ \pm 0.002}$ & $0.67{ \pm 0.002}$ & $0.56{ \pm 0.002}$ & $0.53{ \pm 0.001}$ & $0.59{ \pm 0.001}$ & $0.47{ \pm 0.004}$ & \textbf{3.69} \\
\bottomrule
\end{tabular}
\label{tab:embedding_comparison_f1}
{\footnotesize Model$_{\text{H}}$ refers to Hierarchy Aware embeddings. Column names: "DM" = Diabetes Mellitus, "Periph. Vasc." = Peripheral Vascular Disease, "Neuro." = Neuropathy, "MI" = Myocardial Infarction, "Renal Fail." = Renal Failure, "Readmit" = 30d readmission (MIMIC-IV), * = p$<0.05$}
\end{table*}

\end{document}